\documentclass[10pt]{article} 
\usepackage[preprint]{tmlr}

\usepackage{amsmath,amsfonts,bm}

\def\eqref#1{equation~\ref{#1}}

\def\1{\bm{1}}

\DeclareMathAlphabet{\mathsfit}{\encodingdefault}{\sfdefault}{m}{sl}
\SetMathAlphabet{\mathsfit}{bold}{\encodingdefault}{\sfdefault}{bx}{n}

\usepackage{hyperref}
\usepackage{url}

\usepackage{graphicx}
\graphicspath{{figures/}}

\title{Fisheye Stereo Vision: Depth and Range Error}

\author{\name Leaf Jiang, Matthew Holzel, Bernhard Kaplan, Hsiou-Yuan Liu, Sabyasachi Paul, Karen Rankin, and Piotr Swierczynski  \email leaf@nodarsensor.com \\
      \addr NODAR Inc.}

\def\month{02}  
\def\year{2026} 
\def\openreview{\url{https://openreview.net/forum?id=XXXX}} 

\begin{document}

\maketitle

\begin{abstract}
This study derives analytical expressions for the depth and range error of fisheye stereo vision systems as a function of object distance, specifically accounting for accuracy at large angles.
\end{abstract}

\section{Introduction}
The growing demand for wide-field-of-view sensing in autonomous navigation and robotics has popularized fisheye stereo vision systems. However, traditional depth-mapping models often lose precision in the periphery. This paper addresses this gap by deriving analytical expressions for range error ($\Delta R$) explicitly tailored for fisheye cameras, ensuring mathematical validity even at large incident angles. By characterizing depth discrimination as a function of object distance ($Z$), we establish a rigorous framework to evaluate how wide-angle optics affect spatial perception. 

Prior research in stereo vision has been largely based on the rectilinear pinhole camera model, where depth error ($\Delta Z$) is typically treated as a function of focal length and a fixed horizontal baseline \cite{hartley}. While these models are well-documented for standard fields of view, they fail to account for the radial compression and significant geometric distortion inherent in fisheye optics, particularly at the periphery. In addition, it is not clear how the shortening of the effective baseline length (i.e., the distance between the cameras) further degrades the range error. Here, we present analytic solutions to the range error for both pinhole and fisheye cameras.

In this study, error is a general term that refers to resolution, accuracy, or precision, depending on how the disparity error,  $\Delta d$, is defined.

\begin{itemize}
\item \emph{Resolution.} The smallest detectable difference between two objects or measurements. In lidar, this is equal to the width of the temporal impulse response. For example, for a lidar with $\tau = 4$-ns pulses (or 250 MHz detection bandwidth), the lidar will not be able to differentiate two objects within 60-cm range of each other ($\Delta R = c\tau/2$), or 60-cm range resolution. In stereo vision, this is one pixel. Two objects within one pixel cannot be differentiated. \emph{Resolution} describes the detail and granularity of the measurement.
\item \emph{Accuracy.} The closeness of a measured value to a standard or "true" value (e.g., ground truth surveyed checkpoints). \emph{Accuracy} describes the amount of systematic bias.
\item{Precision.} The closeness with which repeated measurements under unchanged conditions agree with each other, regardless of their relation to the "truth". \emph{Precision} describes the repeatability and consistency of the measurement.
\end{itemize}

In stereo vision systems, \emph{precision} is usually the error metric of interest. In practice, the absolute range calibration can be set to achieve perfect accuracy (whether it remains calibrated is a separate issue), and resolution is seldom discussed because objects of interest typically span more than 1 pixel.

\section{Mathematical Foundation}

The two most common camera models are the pinhole camera model and the fisheye camera model. The pinhole model is a good choice for cameras with fields-of-view up to about 90 degrees, and the fisheye camera model is appropriate for larger fields of view. Both camera models are implemented in \textit{NODAR Hammerhead SDK} \cite{sdk} available at \url{https://www.nodarsensor.com/sdk}.

\begin{figure}[h]
\begin{center}
\includegraphics[width=0.5\linewidth]{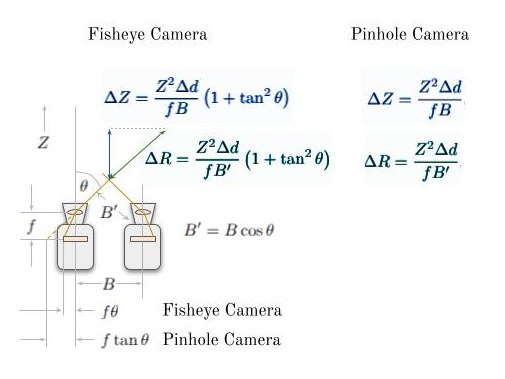}
\end{center}
\caption{Stereo vision cameras. The dashed light rays correspond to the pinhole camera model and the solid light rays correspond to the fisheye camera model. The depth and range errors are shown for both camera models.}
\label{fig:infographic}
\end{figure}

\subsection{Pinhole Camera Model}
The mapping between an incident ray at angle $\theta$ and the radial distance from the center of the image, $r$, for a pinhole camera, as shown in Fig.~\ref{fig:infographic}, is
\begin{equation}
r = f \tan \theta,
\label{eq:pinhole_model}
\end{equation}
where $f$ is the focal length of the lens. For an object at depth Z along the centerline between the cameras with baseline $B$, the disparity, $d$, is
\begin{equation}
\frac{d}{2} = r = f \tan \theta = f \tan \left(\arctan \left(\frac{B/2}{Z}\right) \right) = \frac{fB}{2Z},
\label{eq:similar_triangles}
\end{equation}
where the total disparity, $d$, is the disparity measured on the left camera plus the disparity measured on the right camera In this example, $d = d/2 + d/2$.

The depth error, $\Delta Z$, for a pinhole camera, is obtained by taking the derivative of $d$ in equation~\ref{eq:similar_triangles} with respect to $Z$ and re-writing as
\begin{equation}
\Delta Z = \frac{Z^2 \Delta d}{fB},
\label{eq:dZ_pinhole}
\end{equation}
where $Z$ is the range or distance to the object, $\Delta d$ is the disparity error (the smallest change in disparity that can be reliably measured) in pixels, $f$ is the focal length of the camera in pixels, and $B$ is the baseline, the distance between the two camera centers. The negative sign in equation~\ref{eq:dZ_pinhole} was dropped for simplicity since we often plot the absolute value or square of the error. Equation~\ref{eq:dZ_pinhole} shows that the range error degrades quadratically with the range ($Z$). This means that at greater distances, a small change in disparity corresponds to a very large change in range, making the system less accurate for distant objects.

In technical sensing, depth and range refer to different geometric measurements. Depth (typically denoted as $Z$ in stereo vision) refers to the perpendicular distance from the camera’s focal plane to an object, representing the "depth" within a 3D coordinate system. In contrast, range (commonly used with lidar and denoted as $R$) refers to the direct "radial" distance or line-of-sight vector from the sensor origin to the object point. While stereo vision computes depth through triangulation across an image plane, LiDAR directly measures range via the time-of-flight of a laser pulse. Depth and range are related by the equation
\begin{equation}
\Delta Z = \Delta R \cos\theta.
\label{eq:depth_to_range}
\end{equation}
Substituting equation~\ref{eq:depth_to_range} into equation~\ref{eq:dZ_pinhole} yields the range error for a pinhole camera,
\begin{equation}
\Delta R = \frac{Z^2 \Delta d}{fB'},
\label{eq:dR_pinhole}
\end{equation}
where
\begin{equation}
B' = B\cos\theta
\label{eq:effective_baseline}
\end{equation}
is the effective baseline length at angle $\theta$. When an object is off-axis, near the edge of the camera's field of view, the range error increases because the effective baseline length shortens.

\subsection{Fisheye Camera Model}
The most popular fisheye camera model is the equidistant or f-theta model. The mapping between an incident ray at angle $\theta$ and the radial distance from the center of the image, $r$, for the fisheye model is
\begin{equation}
r = f \theta.
\label{eq:fisheye_model}
\end{equation}
For large angles of incidence ($\sim 1.16 < \theta < \frac{\pi}{2}$), the rays significantly deviate from a pinhole model. The radial distance, $r$, is half the disparity for an object at depth $Z$ along the centerline between the cameras with baseline $B$,
\begin{equation}
\frac{d}{2} = f\theta = f \arctan\left(\frac{B/2}{Z}\right)
\end{equation}
Taking the derivative of $d$ as a function of $Z$ yields
\begin{equation}
\frac{\Delta d}{2} = f \frac{1}{1 + \left(\frac{B/2}{Z}\right)^2} \frac{B/2}{Z^2} \Delta Z
\end{equation}
and the expression can be rewritten as a function of depth error for the fisheye camera
\begin{equation}
\Delta Z = \frac{Z^2 \Delta d}{fB} \left(1 + \tan^2\theta \right)
\label{eq:dZ_fisheye}    
\end{equation}
where $\tan\theta = \frac{B/2}{Z}$. Notice that the depth error for the fisheye camera degrades as $\left(1 + \tan^2\theta \right)$ compared to the depth error of the pinhole camera (equation~\ref{eq:dZ_pinhole}). However, the fisheye camera does not need an infinitely large CMOS sensor, as does the pinhole camera, at large angles.

The range error for the fisheye camera is obtained by substituting \ref{eq:depth_to_range} into \ref{eq:dZ_fisheye} to obtain the following.
\begin{equation}
\Delta R = \frac{Z^2 \Delta d}{fB'} \left(1 + \tan^2\theta \right).
\label{eq:dR_fisheye}
\end{equation}

\section{Fisheye Stereo Performance Analysis}

Consider an 8~MP ($3840 \times 2160$ pixels) high-performance imaging system—such as a 4K security or automotive HDR camera—equipped with a fisheye lens. The sensor features a 2.1~$\mu$m pixel pitch and a 180$^\circ$ horizontal field of view (HFOV) that spans the full 3840-pixel horizontal extent of the CMOS sensor. Applying the projection model in Eq.~\ref{eq:fisheye_model} with $r = 3840/2$ pixels and $\theta = \pi/2$, the effective focal length is determined to be $f = 1222.3$~pixels.

We further evaluate a stereo configuration with a wide baseline of 1.0~m. Historically, ultrawide-baseline stereo vision was considered impractical due to the difficulty of maintaining extrinsic calibration under environmental stressors, such as wind loading or platform vibrations. However, recent advances in real-time autocalibration algorithms \cite{nodar, nodar_patent} have mitigated these stability issues, enabling the use of wide baselines inherent in standard security camera installations.

\begin{figure}[h]
\begin{center}
\includegraphics[width=0.75\linewidth]{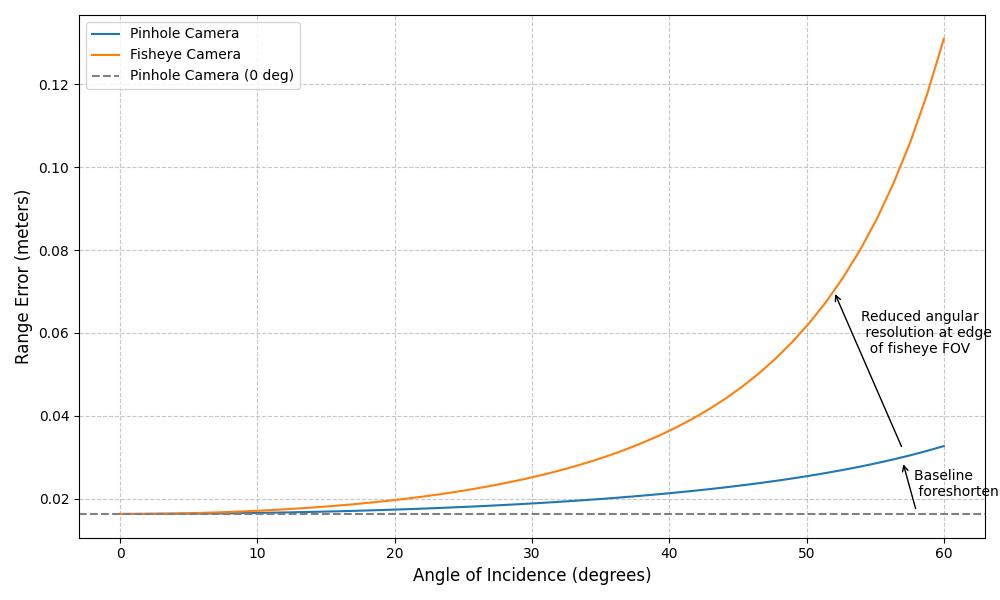}
\end{center}
\caption{Range error for pinhole and fisheye camera plotted as a function of the angle of incidence. The depth, $Z$, is 10~m; focal length, $f$, is 1222.3~pixels; the baseline, $B$, is 1~meter; and the disparity error, $\Delta d$, is 0.2~pixels. This plot was generated from this publicly available notebook \cite{colab}.}
\label{fig:error-vs-theta}
\end{figure}

The range error characteristics for both pinhole and fisheye geometries are illustrated in Fig.~\ref{fig:error-vs-theta} at a depth of $Z = 10$~m, derived from Eq.~\ref{eq:dR_pinhole} and Eq.~\ref{eq:dR_fisheye}. For the fisheye stereo system, the range error remains below 4~cm at a 10-m distance for the incidence angles within $\pm30^\circ$, with the error margins increasing thereafter. Although the pinhole camera’s range error degrades by a factor of $1/\cos\theta$ due to "baseline foreshortening," the fisheye camera exhibits a more pronounced degradation proportional to $1 + \tan^2\theta$. This accelerated error is attributable to the reduction in angular accuracy—specifically the instantaneous field of view (iFOV)—as the projection nears the periphery of the CMOS sensor.

\section{Summary}
This study derives analytical expressions for range and depth error in fisheye stereo vision systems as a function of object distance and angle of incidence. The utility of these derivations is demonstrated through the performance characterization of a 4K fisheye stereo configuration with a 1-m baseline. Our findings reveal a critical divergence in error propagation: whereas pinhole models are primarily constrained by geometric baseline foreshortening, fisheye systems suffer from compounded precision loss at the periphery. This phenomenon is driven by the inherent degradation of angular resolution (iFOV) characteristic of wide-angle optics, necessitating wider baseline distances (enabled by frame-by-frame real-time autocalibration) for wide-field-of-view depth sensing.

\bibliography{main}
\bibliographystyle{tmlr}

\end{document}